\documentclass[11pt]{article}
  \usepackage{amssymb}

  \usepackage[final]{acl}

  \usepackage{times}
  \usepackage{latexsym}
  \usepackage{textalpha}

  \usepackage[T1]{fontenc}
 \usepackage{siunitx}
  \usepackage[utf8]{inputenc}
  \usepackage{amssymb}
\usepackage{subcaption}

  \DeclareFontFamily{LGR}{ptm}{}
  \DeclareFontShape{LGR}{ptm}{m}{n}{<-> ssub * cmr/m/n}{}
  \DeclareFontShape{LGR}{ptm}{b}{n}{<-> ssub * cmr/bx/n}{}
  \DeclareFontShape{LGR}{ptm}{m}{it}{<-> ssub * cmr/m/it}{}
  \DeclareFontShape{LGR}{ptm}{b}{it}{<-> ssub * cmr/bx/it}{}

  \usepackage{microtype}
  \usepackage{inconsolata}

  \usepackage{graphicx}
  \usepackage[most]{tcolorbox}
  \usepackage{booktabs}
  \usepackage{multirow}

  \newtcolorbox{rognote}[1][]{%
    colback=blue!5,
    colframe=blue!60!black,
    coltitle=white,
    fonttitle=\bfseries,
    title=TODOs note,
    boxrule=0.6pt,
    arc=2pt,
    left=6pt, right=6pt, top=4pt, bottom=4pt,
    breakable,
    enhanced,
    #1
  }

\title{Reading or Guessing? Visual Grounding Failures of Vision-Language Models for OCR in Ancient Greek and Arabic Editions}

\author{
Antonia Karamolegkou \and
Nicolas Angleraud \and
Benoît Sagot \and
Thibault Clérice \\
Inria, Paris, France \\
\texttt{\{firstname.lastname\}@inria.fr}
}

\begin{document}
\maketitle

\begin{abstract}
Vision-Language Models (VLMs) used for optical character recognition (OCR) can produce fluent text that is not supported by the page. We study this failure in Ancient Greek and Arabic printed editions by comparing five open-weight VLMs with traditional OCR systems. On real scans, VLMs exhibit errors absent from traditional recognizers, including repetition collapse, markup emission, and off-script generation. Under controlled character perturbations, every VLM moves further from the displayed text than either traditional system in both languages. Token-level analysis shows that similar errors can nevertheless differ in their dependence on the image, while inference-time interventions often reverse direction between Greek and Arabic. These results show that accuracy and fluency alone are insufficient for assessing OCR reliability, provide practical evidence for choosing between general-purpose VLMs and domain-specific recognizers, while also offering an interpretability perspective on how multimodal models balance visual evidence and linguistic priors. 
\end{abstract}

\section{Introduction}
Vision-Language Models (VLMs) are increasingly used for optical character recognition (OCR) and document understanding \citep{bai2025qwen3vltechnicalreport,wei2024gotocr,liu2024ocrbench}. Unlike traditional OCR pipelines, these models generate text autoregressively and may rely on linguistic expectations when the visual evidence is weak or unfamiliar. Recent studies have shown that VLMs can produce plausible but visually unsupported transcriptions, particularly when the linguistic content of an image is perturbed
\citep{shu2025semantic,liang2026visual,he2025seeing,gong2026geometric}. However, evidence comes from modern, high-resource settings, leaving unclear whether the same failures apply to different scripts and historical document processing, a field that uses a lot of OCR.

We examine Ancient Greek and Arabic editions, two challenging settings for OCR systems
\citep{ocr-survey-2020,diao2025ancientscriptimagerecognition}. Ancient Greek critical editions contain polytonic text, mixed scripts, dense layouts, marginal apparatus, and alphabetic numerals. Arabic is cursive and right-to-left, with letter forms that depend on their context. We compare five recent open-weight VLMs with two traditional OCR engines on 90 Greek pages (1{,}966 lines) \citep{angleraud2026icdar} and 140 Arabic pages (3{,}197 lines) \citep{nigst2025openiti}.

We ask three questions: how do VLM and traditional OCR errors differ on real scans; do VLMs reproduce linguistically implausible text when it is clearly visible in the image; and can inference-time interventions reduce these
failures? We combine error analysis, controlled word-order and character perturbations, token-level image-gain measurements, and five inference-time interventions.

Several findings are consistent across the two scripts. Accuracy alone does not separate the two OCR paradigms. The in-domain Kraken recognizer is strongest in both languages, whereas out-of-domain Tesseract is worse than most VLMs. The errors nevertheless differ. Only VLMs exhibit repetition collapse, markup emission, and off-script generation; their errors also affect more lines and their lexical substitutions are more often attested words. Controlled perturbations provide the clearest evidence of language-prior reliance: under character perturbation, every VLM departs further from the displayed text than either traditional recognizer in both languages.

Token-level image-gain analysis adds an distinct qualification. In Greek, the relation between lexical substitutions and perceptual confusions varies by model: olmOCR assigns lower image gain to lexical substitutions,
Qwen3-VL-8B assigns higher image gain, and the difference for Qwen3-VL-2B is not significant. In Arabic, lexical substitutions have descriptively lower median image gain than perceptual confusions for all three analysed models. The intervention results also depend on the setting. Script masking is highly
damaging in Greek but changes little in Arabic; M3ID helps the Greek Qwen models and harms them in Arabic, while VCD shows the reverse pattern. Text-only post-OCR correction improves several Greek systems but degrades nearly all Arabic systems. 

These findings have both practical and analytical implications. Collection-specific OCR can remain more accurate and less prone to open-ended generation failures than general-purpose VLMs. More broadly, grounding cannot be inferred from aggregate accuracy or fluent output alone, and conclusions drawn from one writing system may not transfer to another.

\section{Related Work}

OCR performance remains uneven beyond a small set of
well-represented scripts \citep{liu2024ocrbench,kargaran2026glotocr}.
Historical OCR has therefore largely relied on script- and collection-specific resources and recognizers, including work on polytonic Greek and classical editions
\citep{robertson2017largescale,romanello2021optical,
angleraud2026icdar} and on printed Arabic-script material
\citep{romanov2017important,kiessling2024advances}.
This literature primarily targets recognition accuracy and corpus construction, rather than model comparison and analysis on language priors.

The closest work studies linguistic priors in generative OCR.
\citet{shu2025semantic} show that semantic context can override ambiguous
scene text; \citet{he2025seeing} study hallucination under degraded document
images; and \citet{gong2026geometric} reduce deployment risk through selective
acceptance. Most directly, \citet{liang2026visual} use word- and sentence-level
semantic corruption and find that end-to-end OCR systems degrade more sharply
than traditional pipelines when linguistic support is removed. We reproduce this central behavioural contrast, but extend it in low-resource historical OCR for two scripts and seven models. We further combine character- and word-order perturbations with real-scan error analysis, token-level image-gain measurements, and inference-time interventions. Our focus is therefore not only whether language priors affect OCR, but whether similar prior-consistent errors reflect the same degree of visual grounding across models and scripts.

\section{RQ1: Do VLMs and traditional OCR systems fail differently?}
\label{sec:rq1}

  \begin{table}[t]
  \centering\small\setlength{\tabcolsep}{6pt}
  \begin{tabular}{l rr rr}
  \toprule
   & \multicolumn{2}{c}{\textbf{Ancient Greek}} & \multicolumn{2}{c}{\textbf{Arabic}} \\
  \cmidrule(lr){2-3}\cmidrule(lr){4-5}
  System & CER & WER & CER & WER \\
  \midrule
  \multicolumn{5}{l}{\emph{Traditional recognisers} (CNN--LSTM)}\\

  Tesseract-grc / -ara
  & 7.5 & 23.7 & 24.6 & 77.8 \\
Kraken-CLLG / -OpenITI
  & \textbf{4.1} & \textbf{11.7}
  & \textbf{3.1} & \textbf{18.2} \\
  
  \midrule
  \multicolumn{5}{l}{\emph{Vision-language models}}\\
  LightOnOCR-1B & \textbf{4.1} & 17.8 & 15.4 & 57.1 \\
  DeepSeek-OCR & 6.7 & 26.2 & 20.8 & 66.7 \\
  Qwen3-VL-2B & 8.0 & 34.0 & 17.6 & 60.0 \\
  olmOCR-2-7B & 6.4 & 26.3 & 10.5 & 43.8 \\
  Qwen3-VL-8B & 5.1 & 22.3 & 11.3 & 44.4 \\
  \bottomrule
  \end{tabular}
  \caption{Median error rates (\%). VLMs use greedy decoding ($T{=}0$); traditional baselines are deterministic. Kraken is Kraken-CLLG / Kraken-OpenITI, Tesseract is Tesseract-grc / Tesseract-ara. Best per language bold.}
  \label{tab:rq1_crosslingual}
  \end{table}

We evaluate five open-weight VLMs and two traditional OCR baselines on Ancient Greek and Arabic book scans, reporting Character Error Rate (CER) and Word Error Rate (WER); model and inference details are given in Appendix~\ref{sec:model_justification}. The Greek benchmark contains 1{,}966 lines from 30 sixteenth- to twentieth-century printed editions \citep{angleraud2026icdar}; the Arabic benchmark contains 3{,}197 line crops from 15 held-out nineteenth- and twentieth-century books \citep{nigst2025openiti}. Average line length is similar (8.9 vs.\ 10.2 words), but Greek is scored per page and Arabic per line under different normalization procedures. We therefore report CER and WER within each language and compare system rankings rather than absolute cross-language error rates.

\textbf{Aggregate accuracy does not separate the paradigms.} Table~\ref{tab:rq1_crosslingual} shows variation within both model families. In Greek, traditional and VLM systems interleave: Kraken-CLLG and LightOnOCR-1B both achieve 4.1\% median CER, while Tesseract-grc performs worse than most VLMs. Arabic shows the same pattern at a wider scale: Kraken-OpenITI is best at 3.1\%, whereas Tesseract-ara is worst at 24.6\%. The large gap between Kraken and Tesseract reflects their training data: Kraken is trained in-domain, whereas Tesseract relies mainly on synthetic or modern print.

\textbf{Some VLM errors have no traditional counterpart.} We observe three failure modes only in VLM output: repetition collapse, markup emission, and off-script generation. Repetition affects 1.1\% of olmOCR-2-7B's Greek pages and 0.1--1.5\% of the Arabic outputs of the five VLMs; markup appears in LightOnOCR-1B on both languages and in DeepSeek-OCR on Arabic; and all five VLMs occasionally emit Latin characters within Arabic lines. Neither Kraken nor Tesseract produces any of these behaviours. Such failures can be explained by open-ended generation, whereas the output length and alphabet of traditional recognizers are more tightly constrained by the recognition task. VLM errors are also more widespread: the in-domain Kraken models return substantially more exact lines than the VLMs in both languages. Thus, VLM failures range from small errors across many lines to occasional unbounded generation. Full per-system rates are reported in Appendix~\ref{sec:errors}.

\section{RQ2: Do VLMs remain faithful under linguistic perturbations?}
\label{sec:rq2}

We test for language-prior reliance in two ways: controlled perturbations and token-level image-gain analysis. In the perturbation experiment, each image faithfully renders a modified input, which also serves as the ground truth. A visually grounded system should reproduce the displayed text even when it is linguistically implausible; a system guided by linguistic priors may instead ``repair'' it toward a more plausible form.


\begin{table}[t]
\centering
\small
\setlength{\tabcolsep}{3pt}
\begin{tabular}{
    l
    S[table-format=-1.2]@{}l
    S[table-format=-1.2]@{}l
    S[table-format=2.2]@{}l
    S[table-format=2.2]@{}l
}
\toprule
& \multicolumn{4}{c}{\textbf{Word order}}
& \multicolumn{4}{c}{\textbf{Characters}} \\
\cmidrule(lr){2-5}\cmidrule(lr){6-9}
\textbf{System}
& \multicolumn{2}{c}{Greek}
& \multicolumn{2}{c}{Arabic}
& \multicolumn{2}{c}{Greek}
& \multicolumn{2}{c}{Arabic} \\
\midrule
Tesseract
& -0.01 &    & -0.38 &    
&  6.05 & $^{***}$ &  5.65 &         \\

Kraken
&  0.05 & $^{*}$ &  0.00 &    
&  2.79 & $^{***}$ &  2.53 & $^{***}$ \\
\midrule
LightOn
&  0.20 & $^{**}$ &  0.00 &    
& 12.61 & $^{***}$ & 14.26 & $^{***}$ \\

DeepSeek
&  0.17 &    &  1.65 &    
& 12.62 & $^{***}$ & 23.56 & $^{***}$ \\

Qwen-2B
&  0.35 &    &  1.02 &    
& 42.45 & $^{***}$ & 26.24 & $^{***}$ \\

olmOCR
&  0.62 & $^{***}$ &  0.79 & $^{***}$
& 17.09 & $^{***}$ & 11.32 & $^{***}$ \\

Qwen-8B
&  0.44 & $^{**}$ &  1.24 &    
& 16.85 & $^{***}$ & 10.55 & $^{***}$ \\
\bottomrule
\end{tabular}

\caption{Median paired change in CER (percentage points) from clean to
maximally perturbed text. Positive values indicate higher error against the
perturbed ground truth. Stars denote Holm-adjusted one-sided Wilcoxon tests:
$^{*}p<.05$, $^{**}p<.01$, $^{***}p<.001$.}
\label{tab:rq2_cross}
\end{table}

Table~\ref{tab:rq2_cross} shows the same contrast in both languages. Under word-order perturbations, where each word image remains unchanged, effects are small: VLM paired median changes are at most $+0.62$ percentage points in Greek and $+1.65$ in Arabic, compared with $-0.01$ to $+0.05$ and $-0.38$ to $0.00$, respectively, for the traditional systems. The effect is significant for three of five VLMs in Greek and one of five in Arabic. Character perturbations yield a much larger separation: VLM changes range from $+12.61$ to $+42.45$ points in Greek and from $+10.55$ to $+26.24$ in Arabic, versus $+2.79$ to $+6.05$ and $+2.53$ to $+5.65$ for the traditional systems. Thus, the smallest VLM increase exceeds the largest traditional-system increase in each language. Magnitude and significance differ, however: Tesseract-ara's $+5.65$ change is not significant after Holm correction, whereas Kraken-OpenITI's smaller $+2.53$ change is significant. This pattern is consistent with prior-driven repair, but could also reflect greater sensitivity to unfamiliar character sequences.

\begin{figure}[t]
    \centering
    \includegraphics[width=\columnwidth]{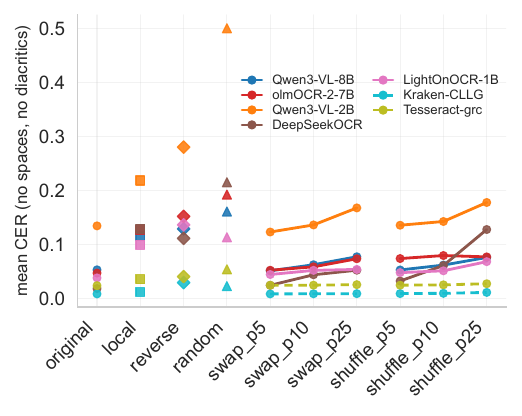}
    \caption{Perturbations averaged across test1 and test2, showing mean CER without diacritics by condition; only proportional swap/shuffle series are connected.}
    \label{fig:rq2}
\end{figure}

We therefore compare each decoded token's image conditioned distribution with an image-free distribution on the Greek benchmark (Appendix~\ref{sec:rq2_imagegain}; Table~\ref{tab:rq2_imagegain}). In Greek, the result is model-specific. For olmOCR-2-7B, fluent lexical substitutions have lower image gain than local
perceptual confusions ($-0.31$ paired within page, $p=.031$), indicating that these errors remain probable without the image. Qwen3-VL-8B shows the opposite pattern ($+0.68$, $p<.001$), remaining image-engaged when substituting whole words, while Qwen3-VL-2B is intermediate and not significant ($+0.10$, $p=.19$). Greek-to-Latin substitutions are highly image-dependent across all three models, suggesting visually triggered confusion rather than image neglect. Figure~\ref{fig:rq2} shows the full perturbation gradient averaged across the two test sets for Ancient Greek. As the degree of perturbation increases, VLM error rises much more rapidly than for the traditional recognizers, with the largest separation under character-level perturbations.

\section{RQ3: Can inference-time interventions restore grounding?}
\label{sec:rq3}
\begin{table}[t]
\centering
\small
\setlength{\tabcolsep}{3pt}
\begin{tabular}{l cc cc}
\toprule
& \multicolumn{2}{c}{\textbf{Greek}}
& \multicolumn{2}{c}{\textbf{Arabic}} \\
\cmidrule(lr){2-3}\cmidrule(lr){4-5}
Intervention & Qwen-2B & Qwen-8B & Qwen-2B & Qwen-8B \\
\midrule
Script mask & $+74.7^{***}$ & $+62.1^{***}$ & $+0.09^{***}$ & $+0.11^{**}$ \\
VCD         & $+0.25$        & $+0.13$        & $-2.95^{***}$ & $-0.50^{***}$ \\
M3ID        & $-1.06^{***}$  & $-1.73^{**}$   & $+7.76^{***}$ & $+2.91^{***}$ \\
LM correct  & $-0.28^{***}$  & $-0.02$        & $+0.87$        & $+2.38^{***}$ \\
\bottomrule
\end{tabular}
\caption{Change in median CER (percentage points) relative to each
intervention's matched baseline; negative values indicate improvement. Greek is scored per page and Arabic per line, so directions rather than magnitudes are comparable across languages. Stars denote unadjusted two-sided paired Wilcoxon signed-rank tests:
$^{*}p<.05$, $^{**}p<.01$, $^{***}p<.001$.
Results for all systems and interventions are reported in
Appendix~\ref{sec:rq3_details}.}
\label{tab:rq3_qwen}
\end{table}

Finally, we ask whether grounding failures can be mitigated at inference time and whether the same interventions transfer across writing systems. We evaluate five methods applied to frozen models: script-constrained decoding and length-based abstention; VCD \citep{leng2024vcd} and M3ID
\citep{favero2024m3id}; and few-shot, text-only post-OCR correction. We exclude methods requiring task-specific training or calibrated controllers
\citep{he2025seeing,gong2026geometric}. Each intervention is compared with its matched baseline using an unadjusted
two-sided paired Wilcoxon signed-rank test. Full results for all systems and interventions, including help/tie/hurt counts,
bootstrap confidence intervals, rank-biserial effect sizes, and adjusted $p$-values, are reported in Appendix~\ref{sec:rq3_details}. Because Greek is scored per page and Arabic per line, we compare effect directions rather than absolute $\Delta$CER across languages.

Most interventions do not transfer consistently across models and scripts.
Script masking damages every VLM in Greek and is highly model-dependent in
Arabic. It changes four Arabic VLMs only slightly, but causes DeepSeek-OCR to
collapse, increasing median CER by $79.2$ points; DeepSeek-OCR is instead the
least affected Greek VLM at $+12.0$ points. M3ID significantly improves both
general-purpose Qwen models in Greek ($-1.06$ and $-1.73$ points) but
significantly worsens them in Arabic ($+7.76$ and $+2.91$). VCD shows the
reverse pattern, with no significant effect on the Greek Qwen models but
significant improvements of $-2.95$ and $-0.50$ points in Arabic. Post-OCR
correction significantly improves four of seven Greek systems but significantly
degrades six of seven Arabic systems. Length abstention improves no median in
either language.

The script-mask results partly reflect the size of the admitted output
vocabulary, but not uniformly across models. For Qwen3-VL-2B, the mask retains
769 Greek-compatible tokens compared with 4{,}197 Arabic-compatible tokens;
olmOCR-2-7B and LightOnOCR-1B have nearly identical counts. This asymmetry is
consistent with their large Greek degradation and small Arabic changes.
DeepSeek-OCR does not follow this pattern: its Arabic mask retains 3{,}426
tokens from a vocabulary of 128{,}827, or 2.7\%, nearly the same proportion
retained for the other models (2.8\%), yet it is the only model that collapses.
In Greek, it admits the largest compatible vocabulary of the five models
(862 tokens) and is the least damaged. The number of admitted tokens therefore
does not explain the outcome on its own, and script constraints require
model-specific validation. Full results are reported in
Appendix~\ref{sec:rq3_details}.

Post-OCR correction repairs text rather than grounding. The corrector receives no image, so improvements show that errors can be recovered from the generated text, not that visual grounding has been restored. The gains are not explained simply by model-family coupling: Qwen3-VL-8B correcting its own Greek output is neutral, whereas the largest improvement is
for Tesseract-grc ($-0.90$ points), which is architecturally unrelated to the corrector. In Arabic, the same correction procedure instead degrades nearly all systems.

\section{Conclusion}
Across Ancient Greek and Arabic historical print, aggregate OCR accuracy does not cleanly separate VLMs from traditional recognizers, but their errors do. VLMs produce more attested substitutions and occasional open-ended generation failures, and under character perturbations every VLM is less faithful to the displayed text than either traditional baseline in both scripts. Token-level image gain varies across models and languages, cautioning against a single explanation for these errors. Several inference-time interventions improve particular model--language combinations, showing promise for mitigation without additional training. 

\section*{Limitations}

Our experiments cover two historical OCR settings, Ancient Greek
critical editions and printed Arabic books. These scripts present
substantially different writing systems and typographic challenges,
but they do not represent the full diversity of historical documents.
The two real-scan benchmarks were independently created for different
scholarly collections and therefore differ in sampling, segmentation,
evaluation unit, and language-specific normalization. Such dataset
heterogeneity is common in multilingual evaluation, where closely
matched historical corpora are rarely available, but it precludes
controlled comparison of absolute error rates across languages. We
therefore compare systems quantitatively within each language and
interpret cross-language agreement only at the level of qualitative
patterns and effect directions.

The perturbation stimuli are rendered rather than naturally scanned.
This provides exact ground truth and isolates responses to changes in
linguistic regularity, but does not reproduce the full range of scan
noise and layout variation found in historical collections. In
Arabic, character permutation can additionally alter contextual glyph
forms. We measure changes in glyph form, ligatures, and rendered
length and find that they do not consistently explain the observed
accuracy loss, although these effects cannot be eliminated entirely.

The mitigation approaches and analysis coverage is model-dependent. Coverage follows the access each analysis requires. \emph{Transcription only}:
  CER/WER, the taxonomy, the perturbation test, length abstention and post-OCR
  correction are post hoc over written output and reach every system in both
  languages. \emph{Per-step distribution}: script-restricted decoding constrains a decoder's next-token distribution, which a CTC recogniser does not have. \emph{Paired image-conditioned and image-free logits}: image gain and contrastive decoding require a second forward pass with the image removed,
  available for the three Qwen-VL-architecture models; LightOnOCR-1B's inference
  path yields no valid image-free comparison and DeepSeek-OCR's architecture is
  unsupported by that harness. 

\section*{Ethical Considerations}

This work uses historical scholarly editions and their transcriptions for OCR evaluation and does not involve human participants or personal data. We cite the CLLG and OpenITI resources from which the evaluation material is derived and follow their documented conditions for research use. Released stimuli and derived annotations are restricted
to material that can be redistributed under the applicable licences.   The Greek material derives from the CLLG corpus of critical editions  \citep{angleraud2026icdar} and the Arabic material from OpenITI \citep{nigst2025openiti}; each is used under its stated terms. Two constraints govern redistribution. The Arabic scans and line transcriptions
  (\texttt{arabic\_print\_data}) carry no declared licence and are therefore
  reserved by default: we use them as research material and release identifiers,
  alignments and metrics rather than page images. The Arabic RQ2 source text is
  the OpenITI RELEASE corpus, licensed CC BY-NC-SA 4.0, so the perturbed Arabic
  stimuli and their derivatives inherit its attribution, share-alike and
  non-commercial terms. The pretrained Kraken recognisers used as in-domain baselines are CC0.

The principal risk concerns downstream use. Linguistically plausible
OCR output can be useful when visual evidence is ambiguous, but
unsupported substitutions may be difficult to detect and can enter
digital editions, search indexes, linguistic corpora, or subsequent
scholarly analyses as if they were faithful transcriptions. This risk
is especially relevant for historical collections, where manual
verification and alternative digital witnesses may be limited. Our
results therefore support collection-specific evaluation, provenance
tracking, and human review for applications in which transcription
fidelity is important. The comparison between model families should
not be read as a general argument against VLM-based OCR: the systems
vary substantially, and linguistic expectations can improve
recognition when they complement rather than replace visual evidence.

To support auditing and reproducibility, we release the controlled
stimuli, evaluation code, and derived error classifications where
licensing permits. AI assistants were used for language editing and
drafting support; all experimental decisions, analyses, and scientific
claims were reviewed and verified by the authors. The experiments
required approximately 100 GPU-hours in total. Detailed compute and
hardware information is reported to make the computational cost of
the study transparent.

\section*{Acknowledgments}

The project ``Corpus Liberatum Linguae Graecae'' was supported by the French National Research Agency (ANR) under the France 2030 grant reference number ``ANR-24-RRII-0002'' operated by the Inria Quadrant Program and is endorsed by the BPI Scribe project.

\bibliography{custom}

\appendix

\
\section{Models and Inference}
\label{sec:model_justification}

\subsection{System selection.}

We compare systems representing two OCR paradigms: traditional recognition-based engines and end-to-end Vision-Language OCR models. 

\paragraph{Traditional OCR systems.}
We compare two traditional recognizers in each language:
\textsc{Tesseract}~v5 \citep{kay2007tesseract} and
\textsc{Kraken}~v5 \citep{kiessling2025kraken}. Unlike the VLMs, both perform
image-conditioned sequence recognition without an autoregressive decoder. For
Greek, we use the released \texttt{grc} Tesseract model
(\textsc{Tesseract-grc}) and the in-domain CLLG Kraken model
(\textsc{Kraken-CLLG}) \citep{angleraud2026icdar}. For Arabic, we use the
released \texttt{ara} Tesseract model (\textsc{Tesseract-ara}) and the
pretrained OpenITI Kraken model (\textsc{Kraken-OpenITI})
\citep{kiessling2022arabicmodel}.

For Greek, \textsc{Tesseract-grc} is trained primarily on synthetic rendered polytonic Greek, whereas \textsc{Kraken-CLLG} is trained on CLLG critical
editions. Real-scan experiments use gold ALTO line polygons, while the perturbation experiments use Kraken's CLLG-trained BLLA segmenter. For Arabic, all systems receive the same gold line crops, so no system performs layout analysis. 
  
\paragraph{Vision-Language Models.}

For VLMs, we evaluate five open-weight models that fall into two categories. The first, \emph{OCR-specialist VLMs}, comprises models post-trained on document-transcription data: \textsc{LightOnOCR-1B} \citep{taghadouini2026lightonocr}, a compact multilingual end-to-end OCR model built on the Pixtral architecture \citep{agrawal2024pixtral}; \textsc{DeepSeek-OCR} \citep{wei2026deepseekocr2}, a 3B-parameter mixture-of-experts model designed around aggressive vision-token compression for efficient document inference; and \textsc{OlmOCR-2-7B} \citep{poznanski2025olmocr}, a 7B OCR-specialist post-trained from a Qwen2.5-VL initialization. The second category consists of \emph{general-purpose VLMs}, \textsc{Qwen3-VL-2B} and \textsc{Qwen3-VL-8B} \citep{bai2025qwen3vltechnicalreport}, two scales of the Qwen3-VL family, which are not OCR-specialized but are widely deployed as document-recognition systems in practice. The lineage relationship between OlmOCR-2 and Qwen2.5-VL lets us isolate the effect of OCR-specific post-training while holding the underlying architecture nearly fixed. We choose these models because they are open-weight, recent, and frequently included in OCR and document-understanding evaluations such as OCRBench, OlmOCR-Bench, and recent multilingual or historical-script OCR studies \citep{liu2024ocrbench,angleraud2026icdar,kargaran2026glotocr}. This selection lets us ask whether visually ungrounded behaviour is specific to large general-purpose VLMs, or whether it also appears in OCR-specialist VLMs that are increasingly used as practical document-recognition systems.

\paragraph{Decoding Setup.}
\label{sec:decoding}

\begin{figure*}[t]
\centering
\includegraphics[width=\linewidth]{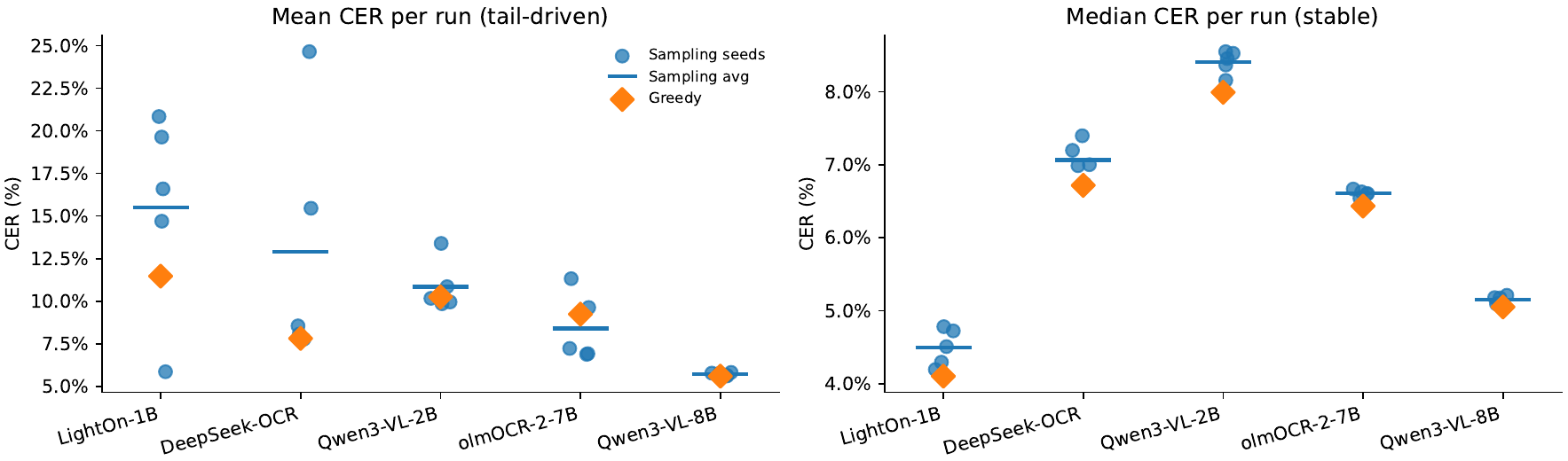}
\caption{Per-seed CER for the five VLMs (dots = the five sampling seeds,
bar = sampling mean, diamond = greedy). Greedy sits within the seed
cluster for every model.}
\label{fig:seed_variance}
\end{figure*}

All main experiments use deterministic greedy decoding via vLLM ($\texttt{temperature}{=}0$, no repetition or frequency penalty, $\texttt{max\_tokens}{=}2048$, image budget $\texttt{max\_pixels}{=}1{,}003{,}520\approx1$\,MP, pages above that budget resized with LANCZOS). This setting reduces generation variance and ensures reproducible error analysis across models and perturbation conditions. Sampling does not systematically improve CER or WER. When performance is summarised by median per-page CER (Table~\ref{tab:greedy_vs_sampling}), the greedy result falls within the range of the sampling seeds for every model. When performance is summarised by mean per-page CER, sampling can increase the error tail for some models, most notably DeepSeek-OCR (sampling mean CER $12.9{\pm}6.5\%$ vs.\ greedy $7.8\%$). Figure~\ref{fig:seed_variance} visualizes the per-seed CER values.

\begin{table}[t]
\centering
\small
\setlength{\tabcolsep}{4pt}
\resizebox{\columnwidth}{!}{%
\begin{tabular}{llrrrr}
\toprule
 & & \multicolumn{2}{c}{CER (\%)} & \multicolumn{2}{c}{WER (\%)} \\
\cmidrule(lr){3-4}\cmidrule(lr){5-6}
Model & Norm. & Greedy & Sampling & Greedy & Sampling \\
\midrule
LightOnOCR-1B & raw & 4.1 & 4.5\,$\pm$0.3 & 17.8 & 18.9\,$\pm$0.3 \\
 & no-diac. & 2.8 & 3.1\,$\pm$0.3 & 11.4 & 12.5\,$\pm$0.4 \\
\addlinespace[2pt]
DeepSeek-OCR & raw & 6.7 & 7.1\,$\pm$0.2 & 26.2 & 27.0\,$\pm$1.0 \\
 & no-diac. & 4.4 & 4.6\,$\pm$0.1 & 15.7 & 16.6\,$\pm$0.7 \\
\addlinespace[2pt]
Qwen3-VL-2B & raw & 8.0 & 8.4\,$\pm$0.2 & 34.0 & 35.4\,$\pm$0.9 \\
 & no-diac. & 4.0 & 4.1\,$\pm$0.1 & 17.6 & 18.0\,$\pm$0.7 \\
\addlinespace[2pt]
OlmOCR-2-7B & raw & 6.4 & 6.6\,$\pm$0.0 & 26.3 & 26.7\,$\pm$0.2 \\
 & no-diac. & 3.4 & 3.6\,$\pm$0.1 & 16.2 & 16.5\,$\pm$0.5 \\
\addlinespace[2pt]
Qwen3-VL-8B & raw & 5.1 & 5.2\,$\pm$0.1 & 22.3 & 22.2\,$\pm$0.3 \\
 & no-diac. & 2.9 & 3.1\,$\pm$0.1 & 13.9 & 14.2\,$\pm$0.3 \\
\bottomrule
\end{tabular}%
}
\caption{Greedy decoding ($T{=}0$) versus stochastic sampling (5 seeds) on the real-scan benchmark. Values are median per-page error rates; sampling reports mean$\pm$std across seed-level medians. \textbf{Norm.} indicates raw text or diacritic-insensitive scoring.}
\label{tab:greedy_vs_sampling}
\end{table}

Per-model prompts follow each model card. \textsc{Qwen3-VL-2B} and \textsc{Qwen3-VL-8B} use  ``\textit{You are an OCR system for [Ancient Greek and Latin] [Arabic] printed scholarly texts. Transcribe the page exactly as it appears without the running header and page numbering. Keep all characters, accents, ligatures, punctuation, and spacing.}''. \textsc{OlmOCR-2-7B} uses the \texttt{OlmOCR} toolkit's \texttt{build\_no\_anchoring\_v4\_yaml\_prompt}; the YAML preamble is stripped from the prediction. \textsc{DeepSeek-OCR} uses the card-default ``\texttt{Free OCR.}'' prompt and \textsc{LightOnOCR-1B} runs image-only with no text prompt. \textsc{Tesseract-grc} runs in v5 with \texttt{-{}-psm 6 -l grc}; \textsc{Kraken-CLLG} uses gold ALTO line polygons on the real benchmark and the CLLG-trained BLLA segmenter with bidirectional reordering on the RQ2 stimuli.

\subsection{Analysis coverage.}

Not every system enters every analysis, and Table~\ref{tab:model_coverage} summarizes the coverage. The aggregate error metrics (CER/WER and the word-level taxonomy) and the perturbation probe cover all seven systems: they require only a predicted transcription and a page image. The token-level image-gain analysis (Appendix~\ref{sec:rq2_imagegain}) and the contrastive decoding interventions (VCD, M3ID) instead require a HuggingFace teacher-forcing path that exposes well-defined per-step image-conditioned logits, available only for the three Qwen-VL--architecture models (Qwen3-VL-2B, Qwen3-VL-8B, and OlmOCR-2-7B); LightOnOCR's inference path did not provide a valid image-conditioned versus image-free logit comparison for this analysis, DeepSeek-OCR's architecture is unsupported by this analysis, and the traditional baselines have no autoregressive decoder logits to query. The decode-time output-space constraints (script-restricted decoding, length abstention) apply to the five VLMs but not to the traditional baselines, which produce a single deterministic transcription per page with no per-step logits to constrain. Post-OCR LM correction, by contrast, is text-only --- it rewrites an existing transcription without access to the page image --- and so applies to all seven systems, including both traditional baselines. A practical consequence is that the token-level grounding signature of RQ2 is established on a single OCR-specialist model, OlmOCR-2-7B. Code and analysis artifacts are available at gitlab link.

\begin{table}[t]
\centering
\small
\setlength{\tabcolsep}{4pt}
\resizebox{\columnwidth}{!}{%
\begin{tabular}{l cc cc ccc}
\toprule
 & \multicolumn{2}{c}{RQ1} & \multicolumn{2}{c}{RQ2} & \multicolumn{3}{c}{RQ3} \\
\cmidrule(lr){2-3} \cmidrule(lr){4-5} \cmidrule(lr){6-8}
System & C/W & Tax & Prt & Gain & Dec & LMc & Con \\
\midrule
Tesseract-grc  & $\bullet$ & $\bullet$ & $\bullet$ & --        & --        & $\bullet$ & --        \\
Kraken-CLLG    & $\bullet$ & $\bullet$ & $\bullet$ & --        & --        & $\bullet$ & --        \\
\midrule
LightOnOCR-1B  & $\bullet$ & $\bullet$ & $\bullet$ & --        & $\bullet$ & $\bullet$ & --        \\
DeepSeek-OCR   & $\bullet$ & $\bullet$ & $\bullet$ & --        & $\bullet$ & $\bullet$ & --        \\
Qwen3-VL-2B    & $\bullet$ & $\bullet$ & $\bullet$ & $\bullet$ & $\bullet$ & $\bullet$ & $\bullet$ \\
olmOCR-2-7B    & $\bullet$ & $\bullet$ & $\bullet$ & $\bullet$ & $\bullet$ & $\bullet$ & $\bullet$ \\
Qwen3-VL-8B    & $\bullet$ & $\bullet$ & $\bullet$ & $\bullet$ & $\bullet$ & $\bullet$ & $\bullet$ \\
\bottomrule
\end{tabular}%
}
\caption{Which system enters which analysis ($\bullet$ included, -- not applicable). \textbf{C/W}: CER/WER; \textbf{Tax}: error taxonomy; \textbf{Prt}: perturbation; \textbf{Gain}: token-level image gain; \textbf{Dec}: script masking and length abstention; \textbf{LMc}: post-OCR LM correction; \textbf{Con}: contrastive decoding. Image gain and contrastive decoding require per-step image-conditioned logits, available only for the three Qwen-VL--architecture models.}
\label{tab:model_coverage}
\end{table}

\section{Error Taxonomy}
\label{sec:errors}

For RQ1 we classify every prediction--reference mismatch into a word-level error taxonomy. The pipeline consists of word-level alignment, fixed
normalization, rule-based category assignment, and lexicon-based identification of attested substitutions. The Greek analysis covers the 90
real scans (17{,}232 reference words; 1{,}966 lines); the Arabic analysis covers 3{,}197 line crops (32{,}892 reference words). Because the two datasets
differ in evaluation unit and normalization, all reported rates should be interpreted within, rather than across, languages.

\paragraph{Alignment.}
Predictions are aligned to their references at the word level using a Levenshtein edit-distance backtrace, producing a sequence of edit operations
(match, substitution, deletion, insertion). Each non-matching operation is assigned exactly one error category, making the taxonomy mutually exclusive and
collectively exhaustive.

\paragraph{Normalization.}
Before alignment we apply fixed normalization to both prediction and reference:
Unicode NFC normalization; removal of reference-only markup
(\texttt{<ref>}, \texttt{<note>}, section markers); rejoining of
line-break hyphenation; normalization of elision apostrophes; isolation of
editorial punctuation; and splitting of digit--letter junctions. Arabic
additionally applies the \texttt{cer\_orth} normalization (unified kaf/yeh
variants, removed tatweel, stripped bidirectional controls), preventing
editorial conventions from contributing to OCR error.

\paragraph{Greek taxonomy.}
Greek uses eight mutually exclusive categories grouped by edit operation:
accent/diacritic, character confusion, cross-script substitution, word
substitution, punctuation, omission, page furniture, and overgeneration
(Table~\ref{tab:error_categories}). Word substitutions combine real-word,
non-word, and segmentation errors, which remain available in the released
analysis files. Repetition runs ($\geq5$ identical tokens) are assigned to
\emph{overgeneration}; inserted or deleted numerals, Latin text, and all-caps
Greek are assigned to \emph{page furniture}.


\begin{table*}[t]
\centering
\small
\setlength{\tabcolsep}{5pt}
\begin{tabular}{p{0.18\textwidth}p{0.43\textwidth}p{0.30\textwidth}}
\toprule
Category & Definition & Examples \\
\midrule
Accent \& diacritic & Same letters; differs only in accent, breathing, case, or final sigma (surface orthography). & αὑτοῖς$\rightarrow$αὐτοῖς; ἐπειδὴ$\rightarrow$ἐπειδή; Ἆρ'$\rightarrow$Ἀρ' \\
\addlinespace
Character confusion & Same-script visual misread of one--two letters ($\leq2$ edits on the bare-letter form). & καὶ$\rightarrow$χαὶ; ὅπως$\rightarrow$δπως; προσβολὴν$\rightarrow$προσδολὴν \\
\addlinespace
Cross-script & A Greek word rendered with Latin look-alike letters (or vice versa). & Παῦλος$\rightarrow$ΠάULO; ὅρκου$\rightarrow$θρsche \\
\addlinespace
Word substitution & A different whole word replaces the target ($>2$ edits, same script). In VLMs this is typically a fluent, in-vocabulary real word; in traditional systems a non-word string. & μισθός$\rightarrow$ἐκούσιος; εἴκοσιν$\rightarrow$ἐκείνων; \emph{(non-word)} φησί$\rightarrow$σπαο; κόλασις$\rightarrow$οττς \\
\addlinespace
Overgeneration & Spurious extra tokens not on the page, including repetition / decoding-loop collapse. & $+$γὰρ; $+$αὐτοῦ; loop: ὅ\,ὅ\,ὅ\dots; X\,X\,X\dots \\
\addlinespace
Omission & A ground-truth word produced by no prediction (skipped or truncated text). & $-$ζητεῖτε; $-$ἠνεωγμένον; $-$ἀπῆλθεν \\
\addlinespace
Page furniture \& ref. & Running heads, page / line / section numbers, and bibliographic Latin that the GT transcription omits. Mostly \emph{faithful reads} of page elements excluded by the transcription standard, not transcription mistakes. & $+$ΕΠΙΣΤΟΛΑΙ; $+$[ΑΙΣΧΙΝΟΥ]; $+$141; $+$PORPHYR \\
\addlinespace
Punctuation & Difference confined to punctuation / quotation / elision marks. & καταφανεῖς·$\rightarrow$καταφανεῖς"; αὐτοῦ;$\rightarrow$αὐτοῦ \\
\bottomrule
\end{tabular}
\caption{Word-level OCR error taxonomy. We align each prediction to its
ground truth with a word-level edit-distance backtrace and assign every
non-matching operation (substitution, insertion, deletion) to exactly one
mutually-exclusive category, so per-model counts partition the errors. Before
alignment we normalise Unicode, unify elision apostrophes, rejoin line-break
hyphenation, and drop GT-only annotation markup ($\langle$ref$\rangle$, section
markers) that the model cannot see. The \emph{Page furniture} category is a
known confound: these tokens inflate the raw error totals of furniture-reading
systems (notably the traditional OCR pipelines) even though most are correct reads of
material the GT excludes. Per-model counts and the finer ten-way breakdown
(real-word vs.\ non-word substitution, word splitting/merging) are in the
released per-model CSVs.}
\label{tab:error_categories}
\end{table*}

  \begin{table*}[t]
  \centering\small\setlength{\tabcolsep}{4pt}
  \begin{tabular}{l rrrrrrrr r}
  \toprule
   & Accent & Char. & Cross- & Word & Punct- & Omis- & Page & Over- & all errors \\
  System & /diacr. & conf. & script & subst. & uation & sion & furn. & gener. & per 1k words \\
  \midrule
  LightOnOCR-1B & 37.1 & 14.8 & 3.5 & 1.9 & 17.6 & 7.1 & 15.7 & 2.3 & 151.8 \\
  DeepSeek-OCR & 53.2 & 18.2 & 0.5 & 4.0 & 8.8 & 2.5 & 10.6 & 2.2 & 277.7 \\
  Qwen3-VL-2B & 57.1 & 18.6 & 0.2 & 6.1 & 6.1 & 4.8 & 5.0 & 2.1 & 392.1 \\
  olmOCR-2-7B & 37.3 & 19.9 & 0.4 & 5.5 & 9.2 & 4.2 & 4.2 & 19.3 & 289.1 \\
  Qwen3-VL-8B & 41.6 & 25.0 & 0.4 & 4.1 & 12.4 & 5.7 & 8.6 & 2.2 & 200.4 \\
  \midrule
  Kraken-CLLG & 15.6 & 12.4 & 0.7 & 8.1 & 15.1 & 10.1 & 31.7 & 6.4 & \textbf{113.2} \\
  Tesseract-grc & 16.9 & 17.4 & 1.1 & 3.6 & 22.6 & 3.1 & 18.8 & 16.5 & 249.6 \\
  \bottomrule
  \end{tabular}
  \caption{Greek word-level error taxonomy on the 90 real scans: each system's
  errors distributed over the eight categories of
  Table~\ref{tab:error_categories}, in \% of that system's total errors, with the
  absolute error burden in the last column. Shares are compositional and must be
  read together with that column: Qwen3-VL-2B's 5.0\% page furniture is a lower
  share than Kraken-CLLG's 31.7\% but a comparable rate (19.6 against 35.9 per
  1{,}000 reference words), because it makes three times as many errors overall.}
  \label{tab:taxonomy_gr}
  \end{table*}
  \begin{table*}[t]
  \centering\small\setlength{\tabcolsep}{3pt}
  \begin{tabular}{l rrrrrrrrrr r}
  \toprule
   & Vocal- & Ortho- & Dot- & Letter- & Cross- & Word & Over- & Omis- & Page & Punct- & all errors \\
  System & isation & graphic & ting & form & script & subst. & gener. & sion & furn. & uation & per 1k words \\
  \midrule
  LightOnOCR-1B & 13.5 & 3.9 & 2.9 & 12.5 & 0.1 & 21.2 & 29.6 & 3.7 & 7.2 & 5.5 & 760.0 \\
  DeepSeek-OCR & 6.5 & 3.3 & 2.1 & 15.3 & 0.2 & 24.9 & 30.6 & 4.0 & 6.9 & 6.3 & 790.1 \\
  Qwen3-VL-2B & 16.2 & 5.3 & 3.3 & 19.5 & 0.2 & 31.3 & 9.8 & 6.1 & 3.6 & 4.8 & 605.6 \\
  olmOCR-2-7B & 11.9 & 9.4 & 2.9 & 17.3 & 0.0 & 27.5 & 16.0 & 3.5 & 3.2 & 8.3 & 460.1 \\
  Qwen3-VL-8B & 17.3 & 7.0 & 2.9 & 14.9 & 0.1 & 28.9 & 13.2 & 5.4 & 4.0 & 6.2 & 493.2 \\
  \midrule
  Kraken-OpenITI & 9.1 & 7.9 & \textbf{11.2} & 15.0 & 0.0 & 25.2 & 5.1 & 13.8 & 3.8 & 8.9 & \textbf{205.8} \\
  Tesseract-ara & 2.6 & 2.4 & 2.5 & 15.0 & 0.0 & 28.8 & 8.3 & 24.7 & 4.4 & 11.2 & 704.5 \\
  \bottomrule
  \end{tabular}
  \caption{Arabic word-level error taxonomy on the 3{,}217 line crops, in \% of
  each system's total errors, with the absolute burden in the last column. Two
  signatures separate the families. \emph{Dotting} --- a misread of the dots
  distinguishing letters that share one skeleton --- is 11.2\% of Kraken-OpenITI's
  errors against 2.1--3.3\% for every VLM: the in-domain recognizer fails below the
  level of the word. \emph{Vocalisation} runs the other way, 11.9--17.3\% for four
  of the five VLMs against 2.6\% for Tesseract-ara: the VLMs supply or delete
  short-vowel marks the print does not determine. \emph{Cross-script} is
  essentially zero for every system, which is why restricting the output to the
  Arabic ranges has nothing to remove (\S\ref{sec:rq3}).}
  \label{tab:taxonomy_ar}
  \end{table*}

\paragraph{Arabic taxonomy.}
The Arabic taxonomy extends the Greek one by replacing the Greek
accent/diacritic category with three script-specific categories:
\emph{vocalization}, \emph{orthographic}, and \emph{dotting}. The remaining
categories are unchanged. Because the Arabic benchmark consists of isolated
line crops, \emph{page furniture} primarily captures inserted or deleted
numerals and non-Arabic tokens rather than running heads. For the traditional
systems, 64--77\% of this category consists of misread digits; for
LightOnOCR-1B, 48\% is emitted markup.

\paragraph{Real-word lexicon.}
To distinguish attested substitutions from non-word strings, Greek uses a
19{,}901-entry lexicon derived from the reference transcriptions of the CLLG
corpus, while Arabic uses a 31{,}243-word external lexicon together with a
lexicon reconstructed from the evaluation references as a robustness check.
A substitution is labelled \emph{real-word} when its normalized bare-letter
form is attested. The Greek lexicon is derived from the same corpus used to
train Kraken-CLLG, making the comparison conservative with respect to
Kraken's domain vocabulary.

\paragraph{Measuring fluency.}
Fluency is measured only for substitutions in which both aligned tokens contain letters, belong to the same script, differ in their bare-letter forms, and are
not repetition artifacts or segmentation fragments. To control for differences in system accuracy, Arabic is additionally evaluated on the 946 reference words
misread by every system. The same restriction is not informative in Greek, where only 13 reference words were misread by all seven systems.


  \begin{table}[t]
  \centering\small\setlength{\tabcolsep}{5pt}
  \begin{tabular}{l rr rr}
  \toprule
   & \multicolumn{2}{c}{Greek} & \multicolumn{2}{c}{Arabic} \\
  \cmidrule(lr){2-3}\cmidrule(lr){4-5}
  System & $n$ & real-word & $n$ & real-word \\
  \midrule
  LightOnOCR-1B & 393 & 16.3 & 9{,}160 & 40.6 \\
  DeepSeek-OCR & 901 & 24.2 & 10{,}970 & 44.1 \\
  Qwen3-VL-2B & 1{,}280 & 24.7 & 10{,}768 & 40.3 \\
  olmOCR-2-7B & 1{,}135 & 32.2 & 7{,}218 & 50.4 \\
  Qwen3-VL-8B & 917 & 32.2 & 7{,}585 & 49.9 \\
  \emph{pooled VLM} & 4{,}626 & \emph{27.2} & 45{,}701 & \emph{44.5} \\
  \midrule
  Kraken & 222 & \textbf{17.1} & 3{,}485 & \textbf{21.9} \\
  Tesseract & 706 & 21.2 & 10{,}721 & 34.3 \\
  \bottomrule
  \end{tabular}
  \caption{Share of misreads whose prediction is an attested form, in \%, over
  every substitution whose letter skeleton changed. The pooled VLM rate exceeds
  both traditional systems in both languages: in Arabic against Kraken-OpenITI
  OR $2.85$ and Tesseract-ara OR $1.53$, in Greek OR $1.81$ and OR $1.38$, all
  $p<.001$. The Greek column counts only substitutions on which two alignments
  agree; see the text for why, and for the difficulty control that leaves the
  Arabic ordering unchanged.}
  \label{tab:fluency}
  \end{table}

\paragraph{Detecting VLM-only failure modes.}
Repetition collapse, markup emission, and off-script output are detected
directly from the raw predictions and therefore do not depend on the alignment
procedure. Repetition collapse is defined as a run of at least five identical
consecutive tokens. Markup emission comprises LaTeX commands, markdown syntax,
HTML tags, and mathematical delimiters. Off-script output is defined as a run
of at least two Latin letters within an Arabic prediction; an analogous Greek
measure is not meaningful because Latin apparatus is frequently present in the
source pages.  \textbf{VLM errors more often involve diacritics.}
Assigning each mismatch to one mutually exclusive category reveals a similar pattern in both languages. In Greek, accents and diacritics account for $37$--$57\%$ of VLM errors,
compared with $16$--$17\%$ for the traditional systems. In Arabic, vocalization errors account for $12$--$17\%$ of errors for four of the five
VLMs, whereas dotting errors are more common for Kraken-OpenITI ($11.2\%$ versus $2$--$3\%$ for the VLMs). Kraken-CLLG's largest Greek category is
instead page furniture excluded from the reference transcription (e.g. page numbers, headers, critical aparatus notes).

  \begin{table}[t]
  \centering\small\setlength{\tabcolsep}{5pt}
  \begin{tabular}{l rr rrr}
  \toprule
   & \multicolumn{2}{c}{Greek (\% of pages)} & \multicolumn{3}{c}{Arabic (\% of lines)} \\
  \cmidrule(lr){2-3}\cmidrule(lr){4-6}
  System & repet. & markup & repet. & markup & off-scr. \\
  \midrule
  LightOnOCR-1B & 0.0 & 6.7 & 1.5 & 2.5 & 2.7 \\
  DeepSeek-OCR & 0.0 & 0.0 & 1.2 & 0.9 & 1.2 \\
  Qwen3-VL-2B & 0.0 & 0.0 & 0.3 & 0.0 & 2.0 \\
  olmOCR-2-7B & 1.1 & 0.0 & 0.1 & 0.0 & 0.1 \\
  Qwen3-VL-8B & 0.0 & 0.0 & 0.1 & 0.0 & 0.4 \\
  \midrule
  Kraken & \textbf{0.0} & \textbf{0.0} & \textbf{0.0} & \textbf{0.0} & \textbf{0.0} \\
  Tesseract & \textbf{0.0} & \textbf{0.0} & \textbf{0.0} & \textbf{0.0} & \textbf{0.0} \\
  \bottomrule
  \end{tabular}
  \caption{Failure modes with no traditional counterpart, measured on raw
  predictions with no alignment step. \emph{repet.}: a run of $\geq5$ identical
  consecutive tokens. \emph{markup}: LaTeX control sequences, math delimiters,
  markdown headers or fences, table rules, HTML tags. \emph{off-scr.}: a run of
  $\geq2$ Latin letters inside Arabic line output; not measurable in Greek, whose
  pages carry Latin apparatus the reference omits. The modes are rare but
  unbounded, and they carry the right tail: LightOnOCR-1B's 49 repeating Arabic
  lines average 6{,}894\% CER and raise its corpus mean from 234\% to 335\%, and
  DeepSeek-OCR's 39 average 1{,}247\%.}
  \label{tab:vlm_only}
  \end{table}

\paragraph{Line-level scoring.}
Exact-line accuracy for Greek is computed by projecting reference line
boundaries onto the page-level predictions using the word alignment, since most
models do not emit reliable line breaks. Inserted words inherit the line index
of the most recently aligned reference word. Arabic is scored directly on the
provided line crops. We report median per-line CER and the proportion of
exactly correct lines.

  \begin{table}[t]
  \centering\small\setlength{\tabcolsep}{5pt}
  \begin{tabular}{l rr rr}
  \toprule
   & \multicolumn{2}{c}{Greek (1{,}966 lines)} & \multicolumn{2}{c}{Arabic (3{,}217 lines)} \\
  \cmidrule(lr){2-3}\cmidrule(lr){4-5}
  System & exact & med.\ CER & exact & med.\ CER \\
  \midrule
  LightOnOCR-1B & 27.3 & 3.5 & 5.8 & 15.4 \\
  DeepSeek-OCR & 14.5 & 5.9 & 3.7 & 20.9 \\
  Qwen3-VL-2B & 6.2 & 8.6 & 4.2 & 17.7 \\
  olmOCR-2-7B & 11.8 & 6.9 & 6.5 & 10.8 \\
  Qwen3-VL-8B & 17.2 & 4.8 & 6.9 & 11.3 \\
  \midrule
  Kraken & \textbf{53.4} & \textbf{0.0} & \textbf{22.2} & \textbf{3.1} \\
  Tesseract & 21.2 & 3.9 & 3.2 & 24.1 \\
  \bottomrule
  \end{tabular}
  \caption{Error concentration: share of lines transcribed exactly, and median
  per-line CER, both in \%. The in-domain recognizer's errors fall on few lines,
  the VLMs' on nearly all of them, which is why a page-level or corpus-level rate
  averages the two into apparent parity. This tracks in-domain training rather than
  architecture: out-of-domain Tesseract patterns with the VLMs in both languages.
  Greek lines are projected from the page-level predictions as described above.
  Arabic per-line scores are available for 2{,}954--3{,}212 of the 3{,}217 lines
  depending on system.}
  \label{tab:exact_lines}
  \end{table}

\section{RQ2 Perturbation and Image-Gain Details}
\label{sec:rq2_details}

\begin{table}[t]
\centering
\small
\setlength{\tabcolsep}{6pt}
\resizebox{\columnwidth}{!}{%
\begin{tabular}{llrrrr}
\toprule
 & & \multicolumn{2}{c}{CER (\%)} & \multicolumn{2}{c}{WER (\%)} \\
\cmidrule(lr){3-4}\cmidrule(lr){5-6}
Model & Norm. & test1 & test2 & test1 & test2 \\
\midrule
Kraken-CLLG    & raw      &  5.6 &  5.6 & 65.3 & 65.3 \\
               & no-diac. &  5.1 &  5.1 & 64.1 & 64.1 \\
\addlinespace[2pt]
Tesseract-grc  & raw      &  6.0 &  6.0 & 62.2 & 62.2 \\
               & no-diac. &  5.5 &  5.5 & 55.9 & 55.9 \\
\midrule
LightOnOCR-1B  & raw      &  7.9 &  7.9 & 67.4 & 67.5 \\
               & no-diac. &  6.3 &  6.3 & 60.9 & 61.0 \\
\addlinespace[2pt]
DeepSeek-OCR   & raw      &  8.0 &  8.0 & 65.9 & 65.5 \\
               & no-diac. &  6.0 &  5.9 & 59.7 & 59.4 \\
\addlinespace[2pt]
Qwen3-VL-2B    & raw      & 19.4 & 18.7 & 91.8 & 91.1 \\
               & no-diac. & 10.3 & 10.2 & 69.1 & 69.1 \\
\addlinespace[2pt]
olmOCR-2-7B    & raw      & 12.0 & 12.0 & 76.1 & 76.1 \\
               & no-diac. &  8.6 &  8.7 & 65.1 & 65.1 \\
\addlinespace[2pt]
Qwen3-VL-8B    & raw      &  9.8 &  9.7 & 72.0 & 71.6 \\
               & no-diac. &  6.6 &  6.6 & 63.0 & 63.0 \\
\bottomrule
\end{tabular}%
}
\caption{RQ2 synthetic benchmark, clean-control (\texttt{original}) condition only. Values are median per-page error rates with jiwer-based scoring, reported for raw text and a diacritic-insensitive variant. test1 and test2 are the word-level and character-level perturbation axes; under \texttt{original}, no perturbation is applied, so differences reflect page-subset differences.}
\label{tab:rq2_synth_original_cer_wer}
\end{table}

We probe reliance on language priors with two complementary methods. The \emph{behavioral} counterfactual-perturbation experiment (Appendix~\ref{sec:rq2_perturb}) asks whether a system transcribes what is on the page or silently ``corrects'' it toward fluent Greek. The \emph{mechanistic} token-level image-gain analysis (Appendix~\ref{sec:rq2_imagegain}) measures, token by token, how much each emitted character actually depends on the image.

\subsection{Counterfactual perturbation stimuli}
\label{sec:rq2_perturb}

\paragraph{Perturbation logic.} A visually grounded reader should transcribe whatever glyphs are on the page, even when they spell nonsense; a reader that leans on a Greek language prior will ``repair'' implausible strings toward real words. We therefore perturb text \emph{before rendering it}, so the rendered image faithfully depicts the perturbed string and the ground truth \emph{is} the perturbed string. Fidelity to the page then shows up as low CER against the perturbed ground truth, whereas prior-driven correction drives CER \emph{up} as the rendered text becomes less word-like.

\paragraph{Sources and rendering.} Greek stimuli are built from the Ancient Greek prose documents that constitute the CLLG synthetic test set. Each text is rendered to A5 pages with \texttt{lualatex} in the GFS~Didot polytonic font (two
 visual styles per document, up to six pages per rendering) following \citet{angleraud2026icdar}; the background-colored document images are from  \citet{Chague_Gallicalbum_2023}. Arabic stimuli are built from the OpenITI RELEASE corpus \citep{nigst2025openiti}, centuries 0300--0700\,AH: 35 books by 35 authors, reassembled from mARkdown into 2{,}422 paragraphs (165{,}881 words) and rendered through the same pipeline in Amiri. The typographic axis differs by language and is part of the design: Greek varies two visual styles per document, Arabic varies ten named typefaces. In both languages per-page ground truth is extracted directly from the compiled PDF with PyMuPDF, so there is no pagination or line-alignment uncertainty between image and reference.

\paragraph{Perturbation axes.} We separate two prior types along orthogonal axes that share a single variant taxonomy:

\begin{itemize}\setlength\itemsep{1pt}
\item \textbf{Test~1 (word-level)} disrupts \emph{word order} while leaving every word intact (a syntactic/positional prior).
\item \textbf{Test~2 (character-level)} disrupts \emph{characters within words} while leaving word boundaries intact (a lexical/orthographic prior).
\end{itemize}

Each axis applies the same operations (acting on words for Test~1, on characters within words for Test~2):
\begin{itemize}\setlength\itemsep{1pt}
\item \texttt{swap\_p\{5,10,25\}}: with probability $p$, swap adjacent units (adjacent words; or two adjacent characters in $p\%$ of words).
\item \texttt{shuffle\_p\{5,10,25\}}: relocate $p\%$ of words; or fully scramble the characters of $p\%$ of words.
\item \texttt{local}: shuffle within non-overlapping windows of three units (words; or characters in every word).
\item \texttt{reverse}: reverse the order of all units (whole-paragraph word order; or characters within each word).
\item \texttt{random}: maximal disruption---per sentence, fully shuffle all words (Test~1) or permute all letters across word slots keeping per-word lengths (Test~2). 
\end{itemize}

\begin{figure*}[t]
\centering
\includegraphics[width=\linewidth]{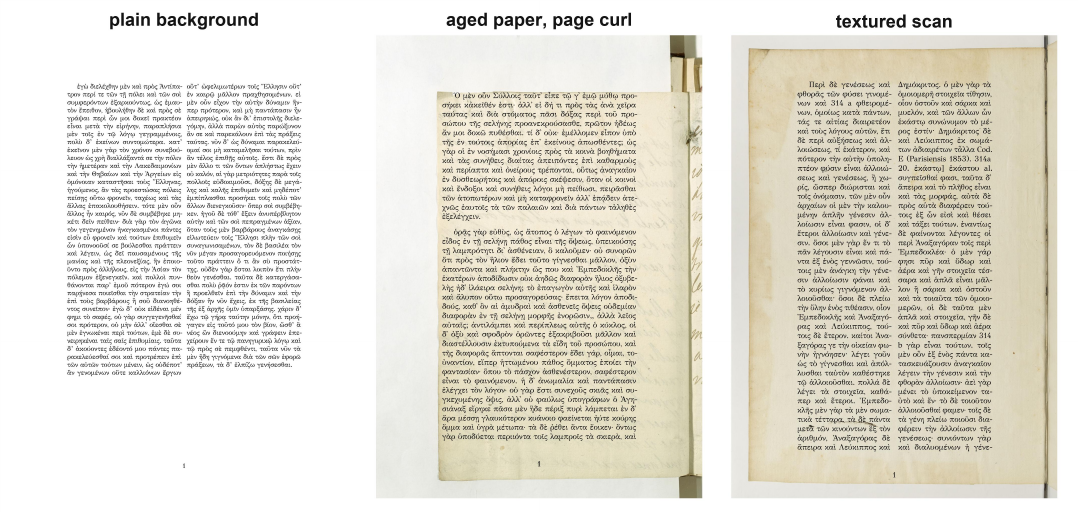}
\caption{Rendered examples from the CLLG synthetic test set used in RQ2. Ancient Greek prose is typeset with \texttt{lualatex} in the GFS~Didot polytonic font and composited onto document-style backgrounds at A5 size, producing controlled page images with varied visual styles. }
\label{fig:rq2_rendering}
\end{figure*}

\begin{figure*}[t]
    \centering

    \begin{subfigure}{\textwidth}
        \centering
        \includegraphics[
            width=\textwidth,
            keepaspectratio
        ]{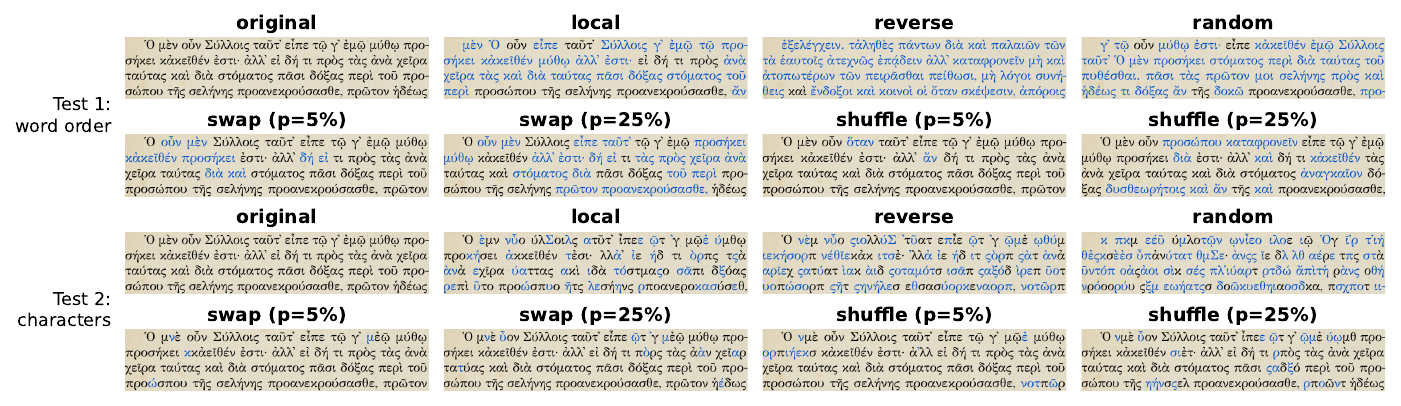}
        \caption{}
        \label{fig:rq2_manipulations_gr}
    \end{subfigure}

    \vspace{0.6em}

    \begin{subfigure}{\textwidth}
        \centering
        \includegraphics[
            width=\textwidth,
            keepaspectratio
        ]{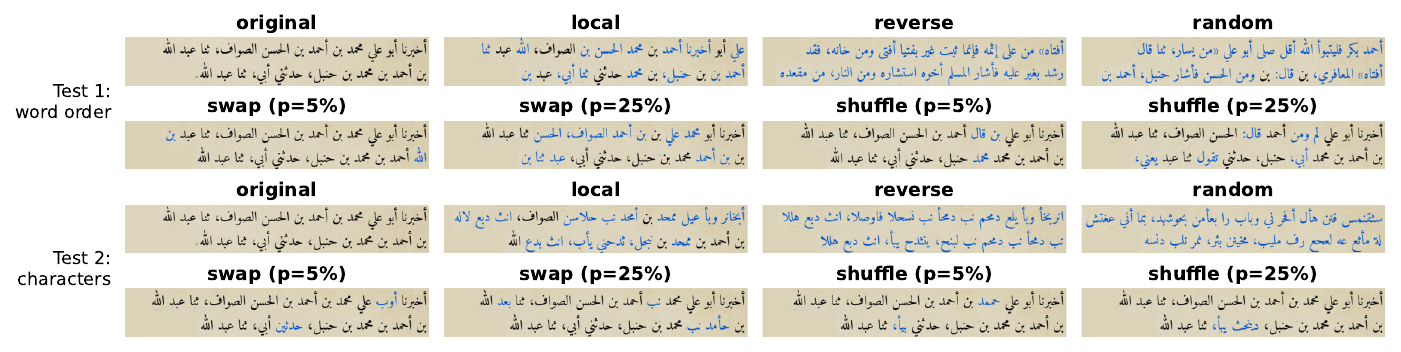}
        \caption{}
        \label{fig:rq2_manipulations_ar}
    \end{subfigure}

    \caption{
    Perturbation conditions illustrated on one passage in
    \textbf{(a)} Ancient Greek and \textbf{(b)} Arabic.
    \textbf{Test~1} is word-level: it changes word order while
    keeping each word intact. \textbf{Test~2} is character-level:
    it applies the same operations within words, preserving word
    boundaries but generally producing non-words.
    \texttt{swap} and \texttt{shuffle} perturb a proportion $p$ of
    units, while \texttt{local}, \texttt{reverse}, and
    \texttt{random} apply stronger sequence disruptions.
    Differences from the original are highlighted in red.
    }
    \label{fig:rq2_manipulations}
\end{figure*}

The \texttt{local}, \texttt{reverse}, and \texttt{random} conditions are
adapted from impossible-language perturbations, which use local shuffles,
reversals, and random shuffles to remove natural word-order structure
\citep{kallini-etal-2024-mission}. The graded \texttt{swap}/\texttt{shuffle}
series follows \citet{liang2026visual}, who perturb OCR inputs to test whether
VLMs rely on language priors rather than visual evidence. Figure~\ref{fig:rq2_rendering} shows example renderings and Figure~\ref{fig:rq2_manipulations} illustrates every condition on a single passage.

\paragraph{Arabic-specific design.}
For Test~2, we permute grapheme units---each base letter together with its combining marks---rather than Unicode code points. This prevents harakat from detaching from their base letters and creating malformed renderings unrelated to the intended perturbation. Additionally, in Arabic, character permutation can change contextual glyph forms as well as letter order. We therefore record, for each stimulus, the proportion of glyphs whose contextual form changes, the change in ligature groups, and the change in
rendered length. Within each perturbation condition, these covariates do not
consistently predict the loss in accuracy.
  
\paragraph{Scoring.}
All systems are scored against the \emph{perturbed} ground truth. For Greek, we report CER after removing spaces and diacritics, isolating letter-level grounding from spacing and accent errors; for Arabic, we use the same orthographically normalised CER as in RQ1. The Greek analysis in Table~\ref{tab} is restricted to single-column pages, since two-column layouts introduce reading-order errors unrelated to grounding. For Arabic, we retain only paragraphs available in every condition within each test: 133 of 135 for Test~1 and 90 of 140 for Test~2. Because the two tests contain slightly different subsets, their clean baselines also differ slightly. We report median paired $\Delta$CER because means are strongly affected by the repetition failures described in \S\ref{sec}. Each $\Delta$ is computed within an example before aggregation, rather than as the difference between marginal medians. Clean CER is not directly comparable between RQ1 and RQ2 because RQ1 evaluates real scans and line crops, whereas RQ2 evaluates rendered single-column controls. Greek is scored per page and Arabic per paragraph, so effect magnitudes should be compared within rather than across languages.

Word-order perturbations have small effects, although the increase is significant for three of five VLMs in Greek and one of five in Arabic. Because each word remains intact, there is relatively little for the model to
``repair.'' Character perturbations produce a much larger contrast: every VLM moves substantially further from the displayed perturbed text than either
traditional recognizer, with no overlap between the two families in either language. This suggests that the strongest prior-driven effect operates at the lexical or orthographic level rather than through sentence-level word order.

\subsection{Token-level image-gain analysis}
\label{sec:rq2_imagegain}

\paragraph{Image gain.} For each model we teacher-force its \emph{own} baseline transcription and, at every generated token $t$, compute
  \begin{align*}
    \mathrm{gain}(t) = \log p(t \mid \text{image},\,\text{prompt})  \\
         - \log p(t \mid \text{prompt}),
  \end{align*}
the log-probability the image adds to the emitted token (an attribution/faithfulness measure). The image-free pass uses the identical prompt with the image removed. We also record the top-1 probability and entropy of the image-conditioned distribution (calibration) and, via a logit lens, the share of probability mass on Greek vs.\ non-Greek vocabulary at intermediate layers.

\paragraph{Token labeling and comparison.}
Predicted characters are aligned to the reference with \texttt{difflib} and
labelled as correct, substituted, or overgenerated; each token inherits the
majority label of its characters. Because correct tokens are dominated by easy,
predictable positions, we compare substitution types with one another rather
than against all correct tokens, restricting the analysis to reference
positions in the page's main script.

Substitutions are divided into three classes: \emph{perceptual}, where the same
base letter differs only in diacritics or orthographic variants;
\emph{cross-script}, where a Latin look-alike replaces a Greek or Arabic
character; and \emph{lexical}, where a same-script substitution produces a
different word. Greek base forms are obtained by removing combining marks.
Arabic uses \texttt{novocal(orth(\,))}, which removes harakat and normalizes
kaf/yeh variants and tatweel; bidirectional controls are removed before
alignment. The analysis covers Qwen3-VL-2B, Qwen3-VL-8B, and olmOCR-2-7B, whose
teacher-forcing implementations expose image-conditioned token logits.

  \begin{table}[t]
  \centering\small\setlength{\tabcolsep}{5pt}
  \begin{tabular}{l rrrr}
  \toprule
   & \multicolumn{4}{c}{median image gain (within-script)} \\
  \cmidrule(lr){2-5}
  Model & correct & percept. & cross-scr. & lexical \\
  \midrule
  \multicolumn{5}{l}{\emph{Greek}} \\
  Qwen3-VL-2B & 1.12 & 1.88 & 4.10 & 2.04 \\
  olmOCR-2-7B & 0.68 & 1.40 & 5.67 & \textbf{0.17} \\
  Qwen3-VL-8B & 0.45 & 1.30 & 5.14 & 1.97 \\
  \midrule
  \multicolumn{5}{l}{\emph{Arabic}} \\
  Qwen3-VL-2B & 3.88 & 2.31 & 0.19 & 1.48 \\
  olmOCR-2-7B & 4.43 & 4.44 & 0.16 & 3.53 \\
  Qwen3-VL-8B & 4.12 & 3.46 & 1.41 & 2.50 \\
  \bottomrule
  \end{tabular}
\caption{Descriptive median image gain by token class. Higher values indicate
stronger image dependence. Absolute values are not comparable across
languages; mechanistic claims use paired within-page comparisons reported in
the text. \textbf{percept.}: local perceptual substitutions;
\textbf{cross-scr.}: substitutions into Latin; \textbf{lexical}: same-script
whole-word substitutions.}
  \label{tab:rq2_imagegain}
  \end{table}

Table~\ref{tab:rq2_imagegain} reports descriptive image-gain medians for both
languages. Because absolute image-gain levels differ substantially across
scripts, we interpret only within-language contrasts and test them using paired
within-page differences.

In Greek, the result is model-specific. Lexical substitutions have lower image
gain than perceptual confusions for olmOCR-2-7B
($\Delta=-0.31$, $p=.031$), while Qwen3-VL-8B shows the opposite pattern
($\Delta=+0.68$, $p<.001$). Qwen3-VL-2B is intermediate and does not differ
significantly ($\Delta=+0.10$, $p=.19$). Arabic does not reproduce this
model-specific split: descriptively, lexical errors have lower median image
gain than perceptual errors for all three models. We therefore treat the
low-image-gain prior-override signature as a Greek finding.

\section{RQ3 Intervention Details}
\label{sec:rq3_details}

All interventions are evaluated on the same benchmarks as in the main text
\citep{angleraud2026icdar,nigst2025openiti} and scored with the same
normalisation. We report median CER as the headline metric because generation
failures produce heavy-tailed error distributions.

\paragraph{Choice of interventions.}
The interventions are chosen to localise where OCR errors can be mitigated,
rather than to construct the strongest possible pipeline. Output constraints
test whether errors arise from the available decoding space; contrastive
decoding tests whether visual dependence can be strengthened at decode time;
and post-OCR correction tests how much can be repaired from text alone.

\paragraph{Engines.}
Script-restricted decoding, length abstention, and post-OCR correction use the
same inference engine as the main baseline. VCD and M3ID require per-step
access to token logits for an additional forward pass, which our serving engine
does not expose. We therefore run each contrastive method and its matched greedy
baseline in HuggingFace Transformers, avoiding comparisons across inference
engines.

\paragraph{Script-restricted decoding.}
At each decoding step, we mask tokens whose decoded strings contain characters
outside the target script, permitted punctuation, or whitespace. Special tokens,
including EOS and role markers, remain available so that generation can
terminate. Masking operates on decoded token strings, allowing byte-level BPE
pieces that realise valid characters. Digits are excluded to prevent copying
page numbers and entering punctuation or length loops.

\paragraph{Length abstention.}
We abstain when the prediction-to-reference length ratio exceeds a fixed
threshold, targeting severe overgeneration. Most pages or lines are unchanged
by this intervention, so we report help, tie, and hurt counts and compute
rank-biserial effect sizes over the affected subset. Length abstention improves
no median in either language.

\paragraph{VCD.}
Visual Contrastive Decoding contrasts logits from the clean image with those
from a diffusion-noised image,
$\ell_{\mathrm{cd}}=(1{+}\alpha)\ell_{\mathrm{clean}}
-\alpha\ell_{\mathrm{noisy}}$, using the adaptive plausibility constraint of
\citet{leng2024vcd}. We use $\alpha=1.0$, $\beta=0.1$, and a noise step of
$500/1000$.

\paragraph{M3ID.}
M3ID replaces VCD's noised-image branch with an image-free contrast,
$\ell_{\mathrm{cd}}=(1{+}w_t)\ell_{\mathrm{image}}
-w_t\ell_{\mathrm{text}}$, where
$w_t=\min(\alpha,e^{\gamma t}-1)$
\citep{favero2024m3id}. We use $\alpha=0.5$ and $\gamma=0.02$ for all three
models with a no-repeat-$3$-gram constraint; for olmOCR-2-7B, we additionally
apply a repetition penalty of $1.15$. Because M3ID requires a custom decoding
loop, we compare it with its own $\alpha=0$ pass under the same decoding
constraints rather than with the serving engine's native greedy output.

\paragraph{Post-OCR LM correction.}
Each baseline transcription is rewritten by Qwen3-VL-8B in text-only mode,
without access to the image, using a fixed three-shot prompt of printed-text
examples in the corresponding language. Improvements therefore show that an
error can be repaired from the transcription alone, not that visual grounding
has been restored.

\begin{table*}[t]
\centering
\scriptsize
\setlength{\tabcolsep}{3.5pt}
\renewcommand{\arraystretch}{1.05}
\begin{tabular}{@{}llrrr rc rr@{}}
\toprule
System & Intervention
& \multicolumn{3}{c}{Pages}
& \multicolumn{2}{c}{$\Delta$ median CER (pp)}
& $r_{\mathrm{rb}}$ & $p$ \\
\cmidrule(lr){3-5}\cmidrule(lr){6-7}
& & Help & Tie & Hurt & $\Delta$ & 95\% CI & & \\
\midrule
\multicolumn{9}{l}{\emph{Ancient Greek; unit = page ($n=90$)}} \\[2pt]

Tesseract-grc
  & Length abstention
  & 0 & 78 & 12
  & $+1.38$ & [$+0.18$, $+2.60$]
  & $+1.00$ & .002 \\
  & Post-OCR correction
  & 68 & 11 & 11
  & $-0.90$ & [$-1.78$, $-0.20$]
  & $-0.83$ & $<$.001 \\

\addlinespace[2pt]
Kraken-CLLG
  & Length abstention
  & 0 & 80 & 10
  & $+0.42$ & [$+0.03$, $+1.44$]
  & $+1.00$ & .005 \\
  & Post-OCR correction
  & 59 & 11 & 20
  & $-0.34$ & [$-0.82$, $-0.01$]
  & $-0.51$ & $<$.001 \\

\addlinespace[2pt]
LightOnOCR-1B
  & Script mask
  & 2 & 0 & 88
  & $+64.5$ & [$+60.0$, $+68.1$]
  & $+0.96$ & $<$.001 \\
  & Length abstention
  & 1 & 80 & 9
  & $+0.32$ & [$+0.02$, $+0.87$]
  & $+0.64$ & .074 \\
  & Post-OCR correction
  & 35 & 24 & 31
  & $-0.33$ & [$-0.78$, $-0.00$]
  & $-0.22$ & .117 \\

\addlinespace[2pt]
DeepSeek-OCR
  & Script mask
  & 3 & 0 & 87
  & $+12.0$ & [$+10.2$, $+14.2$]
  & $+0.99$ & $<$.001 \\
  & Length abstention
  & 0 & 80 & 10
  & $+0.42$ & [$+0.08$, $+1.39$]
  & $+1.00$ & .005 \\
  & Post-OCR correction
  & 56 & 16 & 18
  & $-0.30$ & [$-0.58$, $+0.06$]
  & $-0.55$ & $<$.001 \\

\addlinespace[2pt]
Qwen3-VL-2B
  & Script mask
  & 0 & 0 & 90
  & $+74.7$ & [$+65.6$, $+78.6$]
  & $+1.00$ & $<$.001 \\
  & Length abstention
  & 0 & 80 & 10
  & $+0.52$ & [$+0.00$, $+1.50$]
  & $+1.00$ & .005 \\
  & VCD
  & 42 & 3 & 45
  & $+0.25$ & [$-1.09$, $+1.31$]
  & $+0.03$ & .829 \\
  & M3ID
  & 71 & 0 & 19
  & $-1.06$ & [$-2.65$, $-0.29$]
  & $-0.68$ & $<$.001 \\
  & Post-OCR correction
  & 61 & 3 & 26
  & $-0.28$ & [$-1.04$, $+0.06$]
  & $-0.61$ & $<$.001 \\

\addlinespace[2pt]
olmOCR-2-7B
  & Script mask
  & 1 & 0 & 89
  & $+57.5$ & [$+53.8$, $+62.0$]
  & $+0.96$ & $<$.001 \\
  & Length abstention
  & 1 & 79 & 10
  & $+0.29$ & [$+0.00$, $+0.94$]
  & $+0.67$ & .050 \\
  & VCD
  & 9 & 2 & 79
  & $+1.31$ & [$+0.70$, $+1.99$]
  & $+0.85$ & $<$.001 \\
  & M3ID
  & 33 & 6 & 51
  & $+2.30$ & [$-1.96$, $+5.67$]
  & $+0.22$ & .078 \\
  & Post-OCR correction
  & 42 & 22 & 26
  & $+0.13$ & [$-0.44$, $+0.47$]
  & $-0.25$ & .073 \\

\addlinespace[2pt]
Qwen3-VL-8B
  & Script mask
  & 0 & 0 & 90
  & $+62.1$ & [$+58.5$, $+67.0$]
  & $+1.00$ & $<$.001 \\
  & Length abstention
  & 0 & 80 & 10
  & $+0.32$ & [$+0.02$, $+0.58$]
  & $+1.00$ & .005 \\
  & VCD
  & 30 & 8 & 52
  & $+0.13$ & [$-0.33$, $+0.71$]
  & $+0.22$ & .083 \\
  & M3ID
  & 59 & 1 & 30
  & $-1.73$ & [$-2.73$, $+0.23$]
  & $-0.38$ & .002 \\
  & Post-OCR correction
  & 31 & 26 & 33
  & $-0.02$ & [$-0.29$, $+0.18$]
  & $+0.04$ & .794 \\

\bottomrule
\end{tabular}

\caption{Complete RQ3 results for Ancient Greek (25 cells).
\emph{Help}, \emph{tie}, and \emph{hurt} count pages whose CER decreased,
stayed the same, or increased against each intervention's matched baseline.
$\Delta$ is the change in median CER in percentage points; negative is better.
Confidence intervals are from 10{,}000 paired bootstrap resamples,
$r_{\mathrm{rb}}$ is the matched-pairs rank-biserial correlation over non-zero
differences, and $p$ is an unadjusted two-sided paired Wilcoxon signed-rank
test. Length abstention leaves most pages unchanged, so its effect sizes
describe only the pages it altered. The bootstrap interval estimates the
median paired change, whereas the Wilcoxon test uses the signed ranks of all
non-zero paired differences, so the two summaries need not give the same
significance decision.}
\label{tab:effect_sizes_greek}
\end{table*}

\begin{table*}[t]
\centering
\scriptsize
\setlength{\tabcolsep}{3.5pt}
\renewcommand{\arraystretch}{1.05}
\begin{tabular}{@{}llrrr rc rr@{}}
\toprule
System & Intervention
& \multicolumn{3}{c}{Lines}
& \multicolumn{2}{c}{$\Delta$ median CER (pp)}
& $r_{\mathrm{rb}}$ & $p$ \\
\cmidrule(lr){3-5}\cmidrule(lr){6-7}
& & Help & Tie & Hurt & $\Delta$ & 95\% CI & & \\
\midrule
\multicolumn{9}{l}{\emph{Arabic; unit = line
($n=3{,}217$; Kraken-OpenITI $n=3{,}197$)}} \\[2pt]

Tesseract-ara
  & Length abstention
  & 0 & 3206 & 11
  & $+0.41$ & [$+0.00$, $+0.49$]
  & $+1.00$ & .003 \\
  & Post-OCR correction
  & 629 & 1211 & 1377
  & $+2.24$ & [$+1.67$, $+2.95$]
  & $+0.51$ & $<$.001 \\

\addlinespace[2pt]
Kraken-OpenITI
  & Length abstention
  & 0 & 3190 & 7
  & $+0.00$ & [$+0.00$, $+0.05$]
  & $+1.00$ & .018 \\
  & Post-OCR correction
  & 185 & 714 & 2298
  & $+4.50$ & [$+4.28$, $+4.70$]
  & $+0.93$ & $<$.001 \\

\addlinespace[2pt]
LightOnOCR-1B
  & Script mask
  & 448 & 2379 & 390
  & $-0.10$ & [$-0.51$, $+0.12$]
  & $-0.15$ & $<$.001 \\
  & Length abstention
  & 222 & 2739 & 256
  & $+0.00$ & [$+0.00$, $+0.18$]
  & $-0.31$ & $<$.001 \\
  & Post-OCR correction
  & 791 & 986 & 1440
  & $+2.27$ & [$+1.54$, $+2.66$]
  & $+0.11$ & $<$.001 \\

\addlinespace[2pt]
DeepSeek-OCR
  & Script mask
  & 271 & 125 & 2821
  & $+79.2$ & [$+78.1$, $+80.0$]
  & $+0.82$ & $<$.001 \\
  & Length abstention
  & 170 & 2930 & 117
  & $+0.07$ & [$+0.00$, $+0.27$]
  & $-0.40$ & $<$.001 \\
  & Post-OCR correction
  & 535 & 1206 & 1476
  & $+1.90$ & [$+1.36$, $+2.44$]
  & $+0.34$ & $<$.001 \\

\addlinespace[2pt]
Qwen3-VL-2B
  & Script mask
  & 383 & 2349 & 485
  & $+0.09$ & [$-0.24$, $+0.46$]
  & $+0.14$ & $<$.001 \\
  & Length abstention
  & 43 & 2953 & 221
  & $+0.00$ & [$+0.00$, $+0.00$]
  & $+0.52$ & $<$.001 \\
  & VCD
  & 1706 & 556 & 955
  & $-2.95$ & [$-3.71$, $-2.21$]
  & $-0.46$ & $<$.001 \\
  & M3ID
  & 1112 & 531 & 1574
  & $+7.76$ & [$+6.41$, $+9.18$]
  & $+0.34$ & $<$.001 \\
  & Post-OCR correction
  & 864 & 1101 & 1252
  & $+0.87$ & [$+0.27$, $+1.35$]
  & $+0.03$ & .300 \\

\addlinespace[2pt]
olmOCR-2-7B
  & Script mask
  & 488 & 2175 & 554
  & $-0.19$ & [$-0.41$, $+0.00$]
  & $+0.06$ & .093 \\
  & Length abstention
  & 13 & 3152 & 52
  & $+0.00$ & [$+0.00$, $+0.00$]
  & $+0.33$ & .021 \\
  & VCD
  & 1098 & 960 & 1159
  & $+0.55$ & [$+0.17$, $+0.85$]
  & $+0.03$ & .264 \\
  & Post-OCR correction
  & 207 & 1443 & 1567
  & $+2.29$ & [$+1.97$, $+2.62$]
  & $+0.78$ & $<$.001 \\

\addlinespace[2pt]
Qwen3-VL-8B
  & Script mask
  & 308 & 2521 & 388
  & $+0.11$ & [$-0.07$, $+0.49$]
  & $+0.14$ & .001 \\
  & Length abstention
  & 32 & 3033 & 152
  & $+0.00$ & [$+0.00$, $+0.00$]
  & $+0.45$ & $<$.001 \\
  & VCD
  & 1258 & 990 & 969
  & $-0.50$ & [$-0.91$, $-0.17$]
  & $-0.26$ & $<$.001 \\
  & M3ID
  & 984 & 900 & 1333
  & $+2.91$ & [$+2.41$, $+3.54$]
  & $+0.26$ & $<$.001 \\
  & Post-OCR correction
  & 496 & 1245 & 1476
  & $+2.38$ & [$+2.04$, $+2.82$]
  & $+0.44$ & $<$.001 \\

\bottomrule
\end{tabular}

\caption{Complete RQ3 results for Arabic (24 cells).
\emph{Help}, \emph{tie}, and \emph{hurt} count lines whose CER decreased,
stayed the same, or increased against each intervention's matched baseline.
$\Delta$ is the change in median CER in percentage points; negative is better.
Confidence intervals are from 10{,}000 paired bootstrap resamples,
$r_{\mathrm{rb}}$ is the matched-pairs rank-biserial correlation over non-zero
differences, and $p$ is an unadjusted two-sided paired Wilcoxon signed-rank
test. Five combined script-mask-plus-abstention arms are omitted. Length
abstention leaves most lines unchanged, so its effect sizes describe only the
lines it altered. The bootstrap interval estimates the median paired change,
whereas the Wilcoxon test uses the signed ranks of all non-zero paired
differences, so the two summaries need not give the same significance decision.
olmOCR-2-7B has no M3ID row because that run covered only $800$ of $3{,}217$
lines.}
\label{tab:effect_sizes_arabic}
\end{table*}

\paragraph{Detailed intervention results.}
Tables~\ref{tab:effect_sizes_greek}
and~\ref{tab:effect_sizes_arabic} show that intervention effects depend on both
the model and the script. Script masking damages all five VLMs in Greek,
increasing median CER by $12.0$--$74.7$ points. DeepSeek-OCR is the least
affected Greek model but the most severely affected Arabic one: its median CER
increases by $79.2$ points in Arabic, with 2{,}821 of 3{,}197 lines harmed.
The remaining Arabic VLMs change by at most $0.19$ median CER points.

The script-mask results partly reflect the size of the admitted output
vocabulary, but not uniformly across models. For Qwen3-VL-2B, the mask retains
769 Greek-compatible tokens compared with 4{,}197 Arabic-compatible tokens;
olmOCR-2-7B and LightOnOCR-1B have nearly identical counts. This asymmetry is
consistent with their large Greek degradation and small Arabic changes.
DeepSeek-OCR does not follow this pattern. Its Arabic mask retains 3{,}426
tokens from a vocabulary of 128{,}827, or 2.7\%, nearly the same proportion
retained for the other models (2.8\%), yet it is the only model that collapses.
In Greek, DeepSeek-OCR admits the largest compatible vocabulary of the five
models (862 tokens) and is the least damaged. The number of admitted tokens
therefore does not explain the outcome on its own, and script constraints
require model-specific validation.

The contrastive interventions produce selective gains. M3ID significantly
improves both general-purpose Qwen models in Greek but significantly degrades
them in Arabic. VCD shows the reverse pattern, significantly improving both
Arabic Qwen models while providing no significant effect on either Greek Qwen
model. These effects do not extend to olmOCR-2-7B: VCD significantly worsens
its Greek output and has no significant Arabic effect. Arabic M3ID is not
reported for olmOCR-2-7B because the run completed only 800 of 3{,}197 lines.

Post-OCR correction significantly improves four of seven systems in Greek but
significantly degrades six of seven systems in Arabic; Qwen3-VL-2B is the only
Arabic system without a significant change. Length abstention improves no
median in either language and leaves most pages or lines unchanged. Several
methods therefore show promise for particular model--language combinations,
but none transfers consistently across the evaluated settings.

\end{document}